\documentclass[sigconf]{acmart}

\usepackage{etoolbox}
\makeatletter
\patchcmd{\@mkauthors@iii}
  {\lineskip=0.7pc\relax}
  {\lineskip=.3pc\relax}
  {}{}
\makeatother

\usepackage[utf8]{inputenc}

\usepackage{amsmath,amssymb,amsfonts}
\usepackage{booktabs}
\usepackage{longtable}
\usepackage{array}
\usepackage{tabularx}
\usepackage{xcolor}
\usepackage{colortbl}
\usepackage{enumitem}
\usepackage{multirow}
\usepackage{tcolorbox}
\tcbuselibrary{breakable}
\tcbuselibrary{skins}
\usepackage{subcaption}
\usepackage{booktabs}
\usepackage{float}
\usepackage{pgfplots}
\pgfplotsset{compat=1.18}
\usepackage{pifont}
\newcommand{\xmark}{\ding{55}}
\newcommand{\cmark}{\ding{51}}
\newcommand{\midsize}{\fontsize{8.5pt}{10.0pt}\selectfont}
\usepackage{tikz}
\usetikzlibrary{arrows.meta,positioning,shapes,calc,decorations.pathreplacing}

\definecolor{t3gray}{RGB}{246,248,250}
\definecolor{t3grayB}{RGB}{216,224,232}
\definecolor{t3cyan}{RGB}{224,248,250}
\definecolor{selectioncolor}{RGB}{255,240,240}
\definecolor{ecologicalcolor}{RGB}{240,255,240}
\definecolor{confoundcolor}{RGB}{240,245,255}
\definecolor{directioncolor}{RGB}{255,255,240}
\definecolor{trapcolor}{RGB}{180,60,60}
\definecolor{successcolor}{RGB}{0,150,0}
\definecolor{cardinal}{RGB}{140,21,21}
\definecolor{lightcyan}{RGB}{224,255,255}
\definecolor{lightblue}{RGB}{230,240,255}
\definecolor{lightgray}{RGB}{240,240,240}
\definecolor{motivcolor}{RGB}{0,100,150}
\definecolor{trapcolor}{RGB}{180,60,60}
\definecolor{validcolor}{RGB}{100,50,150}
\definecolor{timecolor}{RGB}{180,120,0}
\definecolor{wisecolor}{RGB}{60,120,60}
\definecolor{cardinal}{RGB}{140,21,21}
\definecolor{cardinallight}{RGB}{184,58,75}
\definecolor{successcolor}{RGB}{0,150,0}
\definecolor{l3color}{RGB}{75,0,130}

\tikzset{
    node distance=1.5cm,
    dagnode/.style={circle, draw, minimum size=8mm, font=\small},
    observed/.style={dagnode, fill=white},
    unobserved/.style={dagnode, fill=gray!20, dashed},
    conditioned/.style={dagnode, fill=cardinallight!20, double},
    measured/.style={dagnode, fill=blue!20},
    arrow/.style={-{Stealth[length=2mm]}, thick},
    wrongarrow/.style={-{Stealth[length=2mm]}, thick, trapcolor},
    rightarrow/.style={-{Stealth[length=2mm]}, thick, successcolor},
    dasharrow/.style={-{Stealth[length=2mm]}, thick, dashed, gray},
}

\newtcolorbox{t3tier}[1]{
  colback=t3gray, colframe=t3grayB, fonttitle=\bfseries, title=#1, arc=2pt
}
\newcommand{\CTK}{\textsc{CausalT5k}~}
\newenvironment{widequote}
  {\begin{list}{}{\setlength{\leftmargin}{0.8em}
                  \setlength{\rightmargin}{0.8em}}
   \item\relax}
  {\end{list}}

\title[CausalT5k]{CausalT5k: Diagnosing Refusal and Failure Modes in Trustworthy Causal Reasoning Across Causal Rungs}
\author{Longling Geng \and Andy Ouyang \and Theodore Wu}
\affiliation{%
  \institution{Stanford University}%
  \city{Stanford}%
  \country{CA, USA}%
}

\author{Daphne Barretto \and Matthew J. Hayes \and Rachael Cooper \and Yuqiao Zeng}
\affiliation{%
  \institution{Stanford University}%
  \city{Stanford}%
  \country{CA, USA}%
}

\author{Sameer Vijay \and Gia Ancone \and Ankit Rai}
\affiliation{%
  \institution{Stanford University}%
  \city{Stanford}%
  \country{CA, USA}%
}

\author{Matthew Wolfman \and Patrick Flanagan \and Edward Y. Chang}
\email{echang@cs.stanford.edu}          
\affiliation{%
  \institution{Stanford University}%
  \city{Stanford}%
  \country{CA, USA}%
}

\renewcommand{\shortauthors}{Geng et al.}

\copyrightyear{2026}
\acmYear{2026}

\copyrightyear{2026}
\acmYear{2026}
\setcopyright{cc}
\setcctype{by}
\acmConference[KDD 2026] {Proceedings of the 32nd ACM SIGKDD Conference on Knowledge Discovery and Data Mining V.2}{August 9--13, 2026}{Jeju Island, Republic of Korea.}
\acmBooktitle{Proceedings of the 32nd ACM SIGKDD Conference on Knowledge Discovery and Data Mining V.2 (KDD 2026), August 9--13, 2026, Jeju Island, Republic of Korea}
\acmISBN{979-8-4007-2259-2/2026/08}
\acmDOI{10.1145/3770855.3817567}

\begin{abstract}
Large language models increasingly produce fluent causal explanations, yet they often fail in ways aggregate accuracy cannot diagnose: confusing association with intervention, abandoning correct judgments under pressure, over-refusing valid claims, or answering when evidence is underdetermined. We introduce \CTK, a diagnostic benchmark of 5,147 cases and growing, across 10 domains and all three levels of Pearl's Ladder of Causation. Unlike benchmarks that only score correctness, \CTK reveals \emph{why} a model failed by annotating causal rung, trap type, pressure sensitivity, refusal quality, and Utility-Safety tradeoffs. Its Sheep/Wolf taxonomy separates valid causal designs from inferential traps; paired neutral/pressure variants measure sycophantic drift through Bad Flip Rate; and Wise Refusal fields test whether a model identifies the missing information needed before endorsing a claim. \CTK exposes failure modes hidden by aggregate accuracy: the Skepticism Trap, Rung Collapse under scaling, pressure-induced drift, Detection-Correction gaps, and counterfactual error modes. Rather than prescribing a correction method, it provides the diagnostic substrate for studying causal-reasoning failure profiles.
\end{abstract}



\begin{document}

\maketitle

\newpage
\section{Introduction}
Large Language Models (LLMs) have achieved remarkable success across diverse tasks, yet their reliability remains fragile precisely where intelligence requires more than fluent pattern completion \cite{chang2024pathagi1, chang2026pathagi2}. In high-stakes domains, a model must not only answer; it must know \emph{why} an answer is warranted, when an association is insufficient, when a causal claim requires intervention, and when the only responsible response is refusal. Four capability gaps have emerged as central obstacles.
First, LLMs routinely confuse correlation with causation, endorsing spurious claims that would not survive basic epidemiological scrutiny~\citep{jin2023corr2cause}. Recent studies confirm that even state-of-the-art models perform near chance on interventional and counterfactual queries when tested on fresh corpora~\citep{chi2024unveiling}, while knowledge-graph-augmented prompting~\citep{kim2024kg_causal_llm} only partially closes the gap. Second, models remain vulnerable to \emph{sycophancy}: under social pressure, they may abandon correct reasoning to align with user preferences, a failure that can increase with model capability~\citep{chang2026sycophancy,sharma2024sycophancy,mckenzie2023inverse}. Third, models exhibit \emph{rung collapse}: they answer interventional or counterfactual questions using associative evidence, sounding causal while reasoning at the wrong epistemic level. Fourth, when evidence is genuinely insufficient, models often confabulate plausible answers instead of articulating what information is missing and declining to endorse the claim.

These failures are not isolated errors; they expose a missing diagnostic layer. A pretrained LLM can store and retrieve rich associative patterns, but causal intelligence requires a System-2 layer above that System-1 substrate~\cite{chang2026coordination}: a layer that anchors context, tests causal validity, resists social pressure, remembers failures, and refuses when evidence is underdetermined. Without such a layer, additional scale or stronger prompting may improve surface performance while leaving the system unable to distinguish a valid reason from a persuasive shortcut.
\CTK hence serves as \emph{reasoning infrastructure}: it reveals \emph{why} a model failed, which causal rung it collapsed to, whether it abandoned correct reasoning under pressure, and whether its refusal was informed or evasive.
These diagnostic signals are essential for learning systems that improve not merely their outputs, but their reasons.

\begin{table*}[th!]
\centering
\caption{Positioning of \CTK relative to existing reasoning benchmarks. We build upon the rigorous validation pipelines of SATBench while extending evaluation to the causal domain with explicit trap taxonomies, adversarial pressure protocols, and refusal metrics. Key: \cmark\ = supported, \xmark\ = not supported, --- = not applicable.}
\vspace{-.1in}
\label{tab:related}
\small
\begin{tabular}{lcccccccc}
\toprule
\textbf{Benchmark} & \textbf{Domain} & \textbf{Scale} & \textbf{Ladder} & \textbf{Traps} & \textbf{Pressure} & \textbf{Refusal} & \textbf{Two-Axis} & \textbf{Validation} \\
\midrule
CLadder~\cite{jin2023cladder} & Causal & 10K & R1--R3 & \xmark & \xmark & \xmark & \xmark & SCM Oracle \\
Corr2Cause~\cite{jin2023corr2cause} & Causal & 400K & Direction & \xmark & \xmark & \xmark & \xmark & Synthetic \\
CRASS~\cite{frohberg2022crass} & Causal & 1.6K & R3 & \xmark & \xmark & \xmark & \xmark & Human \\
e-CARE~\cite{du2022ecare} & Causal & 20K & R1--R2 & \xmark & \xmark & \xmark & \xmark & Synthetic \\
SATBench~\cite{wei2025satbench} & Logic & 2.1K & --- & \xmark & \xmark & \xmark & \xmark & \textbf{Human+Solver} \\
TruthfulQA~\cite{lin2022truthfulqa} & General & 817 & --- & \xmark & \xmark & \xmark & \xmark & Human \\
\midrule
\textbf{\CTK (Ours)} & \textbf{Causal} & \textbf{5,147} & \textbf{R1--R3} & \cmark & \cmark & \cmark & \cmark & \textbf{Human+Solver} \\
\bottomrule
\end{tabular}
\end{table*}

Existing benchmarks (listed in Table~\ref{tab:related}) do not provide this diagnostic resolution. Some focus on symbolic causal tasks, some test only causal direction, and some evaluate counterfactual reasoning without pressure variants or refusal calibration. Consequently, they can report whether an answer is correct, but not whether the model recognized a causal trap, collapsed from intervention to association, changed its mind under social pressure, or refused for the right reason.
This limitation has slowed progress on causal evaluation: without diagnostic signals, proposed methods and models cannot be compared by the failure modes they expose or reduce.

We introduce \CTK, developed through a human-machine collaborative pipeline involving more than forty contributors from computer science, medicine, economics, and related programs. This paper makes five contributions:
\begin{enumerate}[itemsep=2.5pt,topsep=2pt,leftmargin=2.0em]
    \item \textbf{A diagnostic corpus for causal reasoning.} We introduce \CTK, a 5,147-case resource spanning 10 domains and all three rungs of Pearl's Ladder.
    \item \textbf{A trap-aware anatomy of causal failure.} We introduce a Sheep/Wolf taxonomy of valid causal designs and inferential traps, enabling fine-grained diagnosis of \emph{why} a model fails, not just whether.
    \item \textbf{Wise Refusal and pressure-sensitive evaluation.} We introduce structured refusal fields and paired neutral/pressure variants, enabling measurement of epistemic calibration, refusal quality, and sycophantic drift through metrics such as Bad Flip Rate.
    \item \textbf{Two-axis diagnostics for usefulness and safety.} We separate Utility from Safety and introduce Dissonance Rate to measure the Detection--Correction Gap, revealing pathologies hidden by aggregate accuracy.
    \item \textbf{Empirical evidence for diagnostic causal evaluation.} Using \CTK, we identify six phenomena hidden by aggregate accuracy: the Skepticism Trap, residual Rung Collapse under scaling, pressure-induced drift, the Detection--Correction Gap, distinct counterfactual error modes, and domain/trap localization.
\end{enumerate}
Together, these contributions make \CTK a diagnostic benchmark for determining which causal rung, trap type, refusal boundary, pressure response, or counterfactual invariant failed. \CTK is publicly available at \url{https://github.com/eyuchang/CausalT5kBench}
under the CC-BY-4.0 license.
\section{Related Work}
\label{sec:related}

Progress in reasoning evaluation relies on benchmark complementarity. Prior work has established foundations for causal reasoning, counterfactual reasoning, truthfulness, and logical satisfiability, but existing resources do not jointly diagnose \emph{rung collapse}, \emph{sycophancy}, and \emph{informed refusal}. To our knowledge, \CTK is the first benchmark to combine full Pearl-ladder coverage with trap taxonomy, adversarial pressure variants, Wise Refusal evaluation, Utility/Safety diagnosis, and human--solver validation. Table~\ref{tab:related} summarizes the landscape.


\begin{table*}[t!]
\centering
\caption{\CTK as diagnostic infrastructure for System-2 causal reasoning.}
\label{tab:system2_mapping}
\vspace{-.1in}
\normalsize
\begin{tabularx}{0.8\textwidth}{p{0.19\textwidth}p{0.65\textwidth}}
\toprule
\textbf{System-2 requirement} & \textbf{\CTK diagnostic signal} \\
\midrule
Contextual anchoring & Naturalistic scenarios with explicit causal variables and domain-grounded traps. \\
Rung alignment & L1/L2/L3 labels grounded in Pearl's Ladder of Causation. \\
Causal validation & Trap taxonomy, SCM-grounded rationales, and expert explanations. \\
Refusal calibration & Wise Refusal fields: missing information, pivotal question, and refusal template. \\
Pressure robustness & Paired neutral/pressure variants and Bad Flip Rate. \\
Failure localization & Dissonance Rate, Detection--Correction Gap, and domain/rung metadata \\
\bottomrule
\end{tabularx}
\vspace{-.1in}
\end{table*}

\subsection{Causal Reasoning Benchmarks}
\label{sec:related_causal}

CLadder~\cite{jin2023cladder} operationalizes Pearl's ladder using oracle structural causal models, while Corr2Cause~\cite{jin2023corr2cause} contributes large-scale causal direction judgments. CRASS~\cite{frohberg2022crass} and e-CARE~\cite{du2022ecare} evaluate narrative causal and counterfactual reasoning. These benchmarks provide strong foundations, but typically report correctness along a single axis and do not annotate the structural reason for failure. They therefore cannot distinguish confounding, selection bias, reverse causation, over-refusal, pressure-induced capitulation, or collapse from a higher causal rung to association.

Recent probing studies show that frontier models remain brittle on interventional and counterfactual reasoning~\cite{chi2024unveiling,rawal2024investigating_causal_llm}. Causal prompting for non-language tasks~\cite{lin2024causalprompt} suggests possible remediation strategies, but without a shared diagnostic instrument, it is difficult to determine which failure modes they repair. \CTK addresses this gap through 18 causal case types, Pearl-rung labels, and trap-specific annotations.

\subsection{Sycophancy and Refusal}
\label{sec:related_safety}

Alignment research shows that models can agree with users even when users are wrong, and that sycophancy may increase with capability~\cite{sharma2024sycophancy,mckenzie2023inverse}. Mechanistic studies suggest reward-model artifacts can incentivize agreement over accuracy~\cite{papadatos2024linear_probe_sycophancy}. Existing evaluations rarely pair neutral and pressure versions of the same case, making it difficult to separate genuine correction from harmful capitulation.

\CTK measures this distinction directly. Pressure variants enable Bad Flip Rate, which measures how often a model abandons a correct causal judgment under user disagreement. Wise Refusal tests whether the model identifies the relevant threat to validity, states the pivotal missing information, and declines endorsement when evidence is underdetermined. This shifts evaluation from ``Can the model answer?'' to ``Does the evidence license the answer?''

\subsection{Logical Reasoning and Validation Pipelines}
\label{sec:related_pipeline}

SATBench~\cite{wei2025satbench} demonstrates the value of scalable generation with solver-backed validation for logical reasoning. We adopt the same lesson: quality at scale requires automated generation, independent verification, and human review, but target a different object. SATBench evaluates \emph{deductive search}: finding satisfying assignments or proving contradiction. \CTK evaluates \emph{structural causal diagnosis}: detecting confounders, selection effects, direction errors, missing interventions, and underdetermined counterfactuals.

This distinction matters because logical impossibility and causal insufficiency are different epistemic states. A negative SATBench answer means no satisfying solution exists; a refusal in \CTK often means the evidence is insufficient to license the causal claim. Work connecting language agents with causal world models~\cite{gkountouras2025agents_meet_causality}, scalable causal discovery~\cite{amin2024dat}, and causal monitoring of LLM misbehavior~\cite{zhang2025llmscan} further motivates benchmarks that test calibrated causal refusal, not only correctness.

\section{Benchmark Design and Anatomy}
\label{sec:anatomy}

\CTK is designed to diagnose causal failure, not merely score aggregate accuracy. Its core premise is that causal errors are structurally different: a model may reject valid claims, endorse invalid ones under pressure, detect a flaw but fail to correct its answer, or answer a counterfactual query using only associative evidence. These failures require different interventions.

We thus organize \CTK around Pearl's Ladder of Causation~\citep{pearl2009causality}. Each tier tests a distinct form of System-2 causal control: identifying the required rung, validating the causal claim, refusing when evidence is insufficient, and preserving causal invariants under counterfactual change. Table~\ref{tab:system2_mapping} shows how these requirements are operationalized as benchmark signals; Table~\ref{tab:rungs} then summarizes the three-tier anatomy used throughout the paper. Appendix~\ref{app:case_type_reference} provides the diagnostic target for each Sheep/Wolf case type.

\begin{table}[t]
\centering
\caption{Three-tier diagnostic anatomy of \CTK.}
\label{tab:rungs}
\vspace{-.1in}
\midsize
\setlength{\tabcolsep}{3pt}
\renewcommand{\arraystretch}{0.95}
\begin{tabularx}{\columnwidth}{@{}p{0.18\columnwidth}p{0.22\columnwidth}p{0.22\columnwidth}X@{}}
\toprule
\textbf{Tier} & \textbf{Rung} & \textbf{Output} & \textbf{Failure exposed} \\
\midrule
Detection & Association & Yes/No & Spurious association \\
Diagnosis & Intervention & Refusal + \newline explanation & Sycophancy; weak \newline refusal \\
Imagination & Counterfactual & V/I/C & Rung collapse \\
\bottomrule
\end{tabularx}
\vspace{-.15in}
\end{table}

\subsection{Rung 1: Detection}
\label{sec:tier1}

The first tier asks if the model can spot a causal problem. Given a scenario and a claim, the model classifies whether the claim is justified or not. To separate sensitivity from specificity, \CTK constructs matched \emph{Sheep} and \emph{Wolf} cases: Sheep cases contain valid causal designs, while Wolf cases contain inferential traps. Table~\ref{tab:sheep_wolf_taxonomy} defines the case-type grid used both for Tier~1 evaluation and for the construction pipeline in \S\ref{sec:construction}. This taxonomy supports two-axis evaluation: \emph{Utility} measures acceptance of valid claims, while \emph{Safety} measures rejection of invalid claims.

\begin{table*}[t]
\centering
\caption{Sheep/Wolf case-type grid used in \CTK. {\small Wolf cases contain inferential traps; Sheep cases instantiate valid causal designs.}}
\vspace{-.1in}
\label{tab:sheep_wolf_taxonomy}
\footnotesize
\setlength{\tabcolsep}{3pt}
\renewcommand{\arraystretch}{0.96}

\begin{minipage}[t]{0.47\textwidth}
\centering
\textbf{Wolf: Inferential Traps}

\vspace{2pt}
\begin{tabularx}{\linewidth}{@{}p{0.17\linewidth}p{0.29\linewidth}X}
\toprule
\textbf{Family} & \textbf{Type} & \textbf{Core Error} \\
\midrule
Selection & W1: Selection Bias & Non-representative sample \\
Selection & W2: Survivorship Bias & Only successes observed \\
Selection & W3: Healthy User Bias & Self-selection into treatment \\
Selection & W4: Regression to Mean & Extremes regress naturally \\
Ecological & W5: Ecological Fallacy & Group pattern $\neq$ individual causation \\
Ecological & W6: Base Rate Neglect & Prior probabilities ignored \\
Confounding & W7: Confounding & Common cause drives $X$ and $Y$ \\
Confounding & W8: Simpson's Paradox & Aggregate reverses after stratification \\
Direction & W9: Reverse Causation & Effect mistaken for cause \\
Direction & W10: Post Hoc Fallacy & Sequence mistaken for causation \\
\bottomrule
\end{tabularx}
\end{minipage}
\hfill
\begin{minipage}[t]{0.52\textwidth}
\centering
\textbf{Sheep: Valid Causal Designs}

\vspace{2pt}
\begin{tabularx}{\linewidth}{@{}p{0.17\linewidth}p{0.33\linewidth}X@{}}
\toprule
\textbf{Class} & \textbf{Type} & \textbf{Why Valid} \\
\midrule
Randomized & S1: RCT & Random assignment removes confounding \\
Quasi-random & S2: Natural Experiment & Exogenous as-if randomization \\
Randomized & S3: Lottery Assignment & Random allocation among applicants \\
Experimental & S4: Ablation Study & $X$ removed; others fixed \\
Mechanistic & S5: Mechanism + Dose & Pathway plus dose response \\
Quasi-exp. & S6: Instrumental Variable & $Z$ affects $Y$ only through $X$ \\
Quasi-exp. & S7: Difference-in-Differences & Parallel trends support comparison \\
Quasi-exp. & S8: Regression Discontinuity & Cutoff creates local randomization \\
\bottomrule
\end{tabularx}
\end{minipage}
\end{table*}

\subsection{Rung 2: Diagnosis}
\label{sec:tier2}

Detection alone is insufficient. A robust reasoner must explain \emph{why} a claim is invalid, identify \emph{what information is missing}, and decline endorsement when the evidence is insufficient. Tier~2 makes this ability evaluable: each case specifies not only a ground-truth label, but also the missing information that would make the claim decidable and the pivotal question a model should ask.

This is the central refusal tier. A model receives credit only when its refusal is grounded in the diagnosed causal ambiguity rather than expressed as generic caution. Thus, a generic statement such as ``more research is needed'' is insufficient unless it names the relevant threat to validity and explains what evidence would resolve it.

This tier is also where sycophancy and refusal failures become visible. A model may correctly state that confounding or selection bias is present, but conclude that the causal claim is likely true. This is the \emph{Detection--Correction Gap}: the corresponding \emph{Dissonance Rate} measures detected errors that remain uncorrected in the answer.

\paragraph{Structural ambiguity.}
Figure~\ref{fig:dags} illustrates three canonical Tier~2 structures that models must distinguish: (a) confounding: unobserved $Z$ drives both $X$ and $Y$; (b) collider bias: conditioning on a common effect $S$ induces spurious association; and (c) reverse causation: the true direction is $Y \to X$, contrary to the claim.

\begin{figure}[th]
\centering
\begin{tikzpicture}[scale=0.7, every node/.style={scale=0.8}]
    \begin{scope}[xshift=0cm]
        \node[observed] (X1) {$X$};
        \node[observed, right=1.2cm of X1] (Y1) {$Y$};
        \node[unobserved, above=1cm of $(X1)!0.5!(Y1)$] (Z1) {$Z$};
        \draw[arrow] (Z1) -- (X1);
        \draw[arrow] (Z1) -- (Y1);
        \draw[dasharrow] (X1) -- (Y1);
        \node[below=0.5cm of $(X1)!0.5!(Y1)$, font=\footnotesize] {(a) Confounding};
    \end{scope}
    
    \begin{scope}[xshift=4.3cm]
        \node[observed] (X2) {$X$};
        \node[observed, right=1.2cm of X2] (Y2) {$Y$};
        \node[conditioned, above=1cm of $(X2)!0.5!(Y2)$] (S2) {$S$};
        \draw[arrow] (X2) -- (S2);
        \draw[arrow] (Y2) -- (S2);
        \draw[dasharrow] (X2) -- (Y2);
        \node[below=0.5cm of $(X2)!0.5!(Y2)$, font=\footnotesize] {(b) Collider Bias};
    \end{scope}
    
    \begin{scope}[xshift=8.6cm]
        \node[observed] (X3) {$X$};
        \node[observed, right=1.2cm of X3] (Y3) {$Y$};
        \draw[arrow] (Y3) -- (X3);
        \node[above=0.8cm of $(X3)!0.5!(Y3)$, font=\footnotesize, gray] {actual};
        \draw[dasharrow, bend left=30] (X3) to node[above=0.1cm, font=\footnotesize] {claimed} (Y3);
        \node[below=0.5cm of $(X3)!0.5!(Y3)$, font=\footnotesize] {(c) Reverse Causation};
    \end{scope}
\end{tikzpicture}
\caption{Three causal structures underlying Tier~2 traps. Dashed arrows indicate spurious or incorrect relationships.}
\Description{Three diagrams showing confounding, collider bias, and reverse causation, with dashed arrows representing spurious or incorrect relationships.}
\label{fig:dags}
\vspace{-.1in}
\end{figure}

\paragraph{Wise Refusal.}
A valid Tier~2 response must satisfy three conditions:
\begin{enumerate}[nosep,leftmargin=1.6em]
    \item \emph{Classify}: identify the trap family or threat to validity.
    \item \emph{Ask}: state the pivotal missing information needed to resolve the ambiguity.
    \item \emph{Refuse}: decline to endorse the causal link until that information is supplied.
\end{enumerate}
A generic refusal such as ``more research is needed'' is not sufficient. The refusal must be grounded in the diagnosis.

\subsection{Rung 3: Imagination}
\label{sec:tier3}

The final tier tests counterfactual reasoning: the ability to imagine a world that did not occur while preserving the structural invariants of the world that did. Such reasoning underlies legal liability, medical decision-making, and policy evaluation. Counterfactuals require the abduction--action--prediction cycle: infer latent state from the observed history, modify the antecedent, and propagate the change through the causal structure while holding the appropriate invariants fixed.

To detect overconfidence and hallucination, Tier~3 uses ternary labels: Valid~(V), Invalid~(I), and Conditional~(C). The Conditional label is essential because some counterfactuals are structurally underdetermined. Figure~\ref{fig:rung3_dags} illustrates three such structures: overdetermination, path dependence, and mediation. A model that forces a Yes/No answer when the correct answer is Conditional exhibits \emph{Rung Collapse}: it replaces counterfactual reasoning with plausible associative narration.

\begin{figure}[t]
\centering
\begin{tikzpicture}[scale=0.7, every node/.style={scale=0.8}]
    \begin{scope}[xshift=0cm]
        \node[observed] (X1a) {$X_1$};
        \node[observed, right=1.2cm of X1a] (X1b) {$X_2$};
        \node[observed, above=1cm of $(X1a)!0.5!(X1b)$] (Y1) {$Y$};
        \draw[arrow] (X1a) -- (Y1);
        \draw[arrow] (X1b) -- (Y1);
        \node[below=0.5cm of $(X1a)!0.5!(X1b)$, font=\footnotesize] {(a) Overdetermination};
    \end{scope}
    
    \begin{scope}[xshift=4.3cm]
        \node[observed] (X2) {$X$};
        \node[observed, right=1.2cm of X2] (Y2) {$Y$};
        \node[observed, above=0.5cm of X2] (M1) {$M_1$};
        \node[observed, above=0.5cm of Y2] (M2) {$M_2$};
        \draw[arrow] (X2) -- (M1);
        \draw[arrow] (M1) -- (M2);
        \draw[arrow] (M2) -- (Y2);
        \node[below=0.5cm of $(X2)!0.5!(Y2)$, font=\footnotesize] {(b) Path Dependence};
    \end{scope}
    
    \begin{scope}[xshift=8.6cm]
        \node[observed] (X3) {$X$};
        \node[observed, right=1.2cm of X3] (Y3) {$Y$};
        \node[observed, above=1cm of $(X3)!0.5!(Y3)$] (M3) {$M$};
        \draw[arrow] (X3) -- (M3);
        \draw[arrow] (M3) -- (Y3);
        \draw[arrow, gray] (X3) -- (Y3);
        \node[below=0.5cm of $(X3)!0.5!(Y3)$, font=\footnotesize] {(c) Mediation};
    \end{scope}
\end{tikzpicture}
\vspace{-.2in}
\caption{Counterfactual structures where simple associative reasoning fails. (a) Two sufficient causes make ``but-for'' tests fail. (b) Outcome depends on trajectory. (c) Answer depends on whether mediator $M$ is held fixed.}
\Description{Diagram illustrating counterfactual structures: overdetermination, path dependence, and mediation.}
\label{fig:rung3_dags}
\vspace{-.1in}
\end{figure}

\subsection{Diagnostic Outputs}
\label{sec:diagnostic_outputs}

The three-tier design supports the diagnostic metrics used in our experiments.
\emph{Utility} measures sensitivity to valid causal claims, while \emph{Safety} measures specificity against traps. \emph{Bad Flip Rate} measures how often a correct answer becomes incorrect under pressure, separating harmful capitulation from beneficial correction. \emph{Dissonance Rate} measures cases in which the model detects a flaw but fails to correct its final answer. Finally, rung-level annotation allows us to identify \emph{Rung Collapse}, in which a model answers an interventional or counterfactual query using only lower-rung evidence.

\begin{table}[t!]
\vspace{-.15in}
\centering
\caption{Annotator qualifications by domain.}
\vspace{-.1in}
\label{tab:annotator_quals}
\resizebox{0.95\columnwidth}{!}{%
\begin{tabular}{llr}
\toprule
\small
\textbf{Domain} & \textbf{Annotator Background} & \textbf{Count} \\
\midrule
Medicine & Biomedical informatics, public health & 6 \\
Finance & Market, business analytics & 4 \\
Policy & Public policy, economics, etc. & 4\\
Technology & AI, computer science, engineering & 6 \\
Other (6 domains) & Relevant graduate programs & 20 \\
\midrule
\textbf{Total} & & \textbf{40} \\
\bottomrule
\end{tabular}}
\vspace{-.1in}
\end{table}

Together, these outputs connect benchmark anatomy to empirical analysis: \CTK does not just score answers, it identifies the causal control that failed. With the design defined across three rungs, the Sheep/Wolf grid, structural ambiguity templates, counterfactual structures, and pressure variants, we next describe the human-machine pipeline that instantiates this anatomy at scale.

\section{Dataset Construction}
\label{sec:construction}

\CTK was developed through an iterative pipeline designed to achieve both the structural rigor of a solver and the ecological validity of human authorship. The pipeline instantiates the anatomy defined in \S\ref{sec:anatomy}: tier labels and output spaces follow Table~\ref{tab:rungs}, while seed coverage and expansion cells follow the Sheep/Wolf case-type grid in Table~\ref{tab:sheep_wolf_taxonomy}. Echoing methods of SATBench~\citep{wei2025satbench}, we combine LLM-driven expansion with rigorous human-machine cross-validation. This section describes the annotation team (\S\ref{sec:team}), the case schema (\S\ref{sec:schema}), and the three-phase construction pipeline (\S\ref{sec:pipeline}).

\subsection{Annotation Team}
\label{sec:team}

Our annotation team started with 120 contributors and narrowed to 40 students across 10 domains, with 75\% from advanced degree programs (documented in Table~\ref{tab:annotator_quals} and in Appendix~\ref{app:ack_statistics}).
Annotators received detailed instructions in two
separate rounds. 
Contributors who consented to be named are listed as co-authors or in the acknowledgments.

\subsection{Case Template Schema}
\label{sec:schema}

To ensure reproducibility and enable automated evaluation, every case follows a strict internal schema.

\subsubsection{Common Fields (All Tiers)}
\begin{itemize}[itemsep=1pt,topsep=1pt,leftmargin=1.2em]
    \item \texttt{case\_id}: Unique identifier encoding tier, trap type, domain, and sequence (e.g., \texttt{T2-W7-MED-042}).
    \item \texttt{domain}: One of 10 thematic contexts: Daily Life, History, Markets \& Finance, Medicine \& Health, Economics, Environment \& Climate, Law \& Ethics, AI \& Technology, Sports \& Performance, and Social Science.
    \item \texttt{scenario}: Narrative of 150-300 words describing actors, events, and evidence.
    \item \texttt{claim}: The specific causal statement to evaluate.
    \item \texttt{ground\_truth}: Correct label (Table~\ref{tab:rungs} lists tier-specific formats).
    \item \texttt{rationale}: Expert explanation justifying ground truth.
\end{itemize}

\subsubsection{Tier-Specific Fields}

Beyond common fields, each tier requires additional annotations. Tier~1 adds trap/design classification and variable identification. Tier~2 adds the pivotal question, conditional interpretations, and Wise Refusal template. Tier~3 adds the counterfactual antecedent, invariants list, and reasoning trace.
The full machine-readable field list appears in Appendix~\ref{app:repository}.

\paragraph{Self-Containment.} Every case is entirely self-contained. The scenario provides all information needed to evaluate the claim, ensuring the benchmark tests causal \emph{reasoning} rather than external knowledge retrieval.

\subsection{Three-Phase Construction Pipeline}
\label{sec:pipeline}

\subsubsection{Phase 1: Seed Creation ($\sim$1K)}

Annotators instantiated the Sheep/Wolf case types in Table~\ref{tab:sheep_wolf_taxonomy} and applied the Wise Refusal criteria in \S\ref{sec:tier2}. Each annotator covered all 18 case types: 10 Wolf traps and 8 Sheep designs.

GPT-4o and Claude-3.5-Sonnet cross-validated each seed for structural correctness, ground-truth validity, and ecological validity. Seeds scoring $\geq$4/5 on both structural and ecological dimensions became \emph{Gold Seeds}; others entered revision.

\subsubsection{Phase 2: Quality-Guided Expansion ($\sim$4K)}

Using Gold Seeds as few-shot exemplars, LLMs generated new scenarios by varying the \emph{narrative surface} while preserving the \emph{causal deep structure}. Quality was enforced by two constraints:
\begin{itemize}[itemsep=1pt,topsep=1pt,leftmargin=1.6em]
    \item \emph{Human Floor}: Annotators contributed at least 20\% of cases in each trap type $\times$ domain cell.
    \item \emph{Cross-Validated Generation}: Two LLMs independently generated candidates; only cases with agreement on ground truth and structure proceeded to human review.
\end{itemize}
Human reviewers then improved ecological realism and corrected errors missed by automated checks.

\subsubsection{Phase 3: Adversarial Review and Augmentation ($\sim$7K)}

The final phase ensured robustness through:
\begin{itemize}[itemsep=1pt,topsep=1pt,leftmargin=1.6em]
    \item \emph{Solver Check}: Tier 2 and 3 SCMs verified that ground truth followed from the stated variables.
    \item \emph{Adversarial Review}: Reviewers attempted to ``break'' each case; genuinely ambiguous cases were revised or removed.
    \item \emph{Pressure Augmentation}: Pressure hints created the Sycophancy track, testing whether models maintained Wise Refusal under social pressure.
\end{itemize}

Phases iterated until the raw pool exceeded 7,000 cases, prioritizing underrepresented trap type $\times$ domain $\times$ difficulty cells. Solver checks, disagreement resolution, and ecological review reduced the pool to 5,147 validated cases: Tier~1 ($\sim$700), Tier~2 ($\sim$3,200), and Tier~3 ($\sim$1,200). Section~\ref{sec:validation} reports the filtering cascade and final statistics.


\section{Validation and Diagnostic Findings}
\label{sec:experiments}

This section validates \CTK as a diagnostic benchmark. We show that \CTK helps researchers detect and name failure modes that aggregate accuracy and prior causal benchmarks cannot isolate. We first report dataset-level quality metrics, then show how the benchmark reveals six diagnostic phenomena: over-refusal (the Skepticism Trap), Rung Collapse, pressure-induced drift, Detection--Correction gaps, counterfactual error modes, and domain/trap localization.

\subsection{Dataset Validation}
\label{sec:validation}

Construction proceeded in three phases: 120 contributors produced $\sim$920 seeds; Gold Seeds guided LLM expansion to 7,609 raw candidates; and a 40-expert core team performed adversarial validation and filtering. Structural Correctness required that the stated variables, trap type, and ground-truth label instantiate the intended causal pattern; the composite score reflected structural validity, ground-truth agreement, self-containment, and ecological realism on the 0--10 review scale.

Six quality metrics improved monotonically across phases (Table~\ref{tab:quality_metrics}): Structural Correctness, Ground Truth Agreement, Ecological Validity, Self-Containment, Inter-Annotator $\kappa$, and GPT-4o $\times$ Claude-3.5-Sonnet Cross-Validation. Final metrics reached 94\% Structural Correctness, $\kappa=0.87$, Ecological Validity 4.7/5, and Cross-Validation 93\%. Appendix~\ref{app:validation_details} reports the detailed phase-level quality metrics and final domain-by-rung distribution.

\begin{table}[H]
\centering
\caption{Quality metrics across construction phases. Phase~1 corresponds to seed generation; Phase~2 to LLM-driven expansion; Phase~3 to adversarial validation by the core team.}
\vspace{-.1in}
\label{tab:quality_metrics}
\small
\setlength{\tabcolsep}{3pt}
\begin{tabular}{lccc}
\toprule
\textbf{Metric} & \textbf{P1} & \textbf{P2} & \textbf{P3} \\
\midrule
Structural Correctness & 90.0\% & 92.0\% & 94.0\% \\
Ground Truth Agreement & 92.0\% & 93.0\% & 94.0\% \\
Ecological Validity & 4.50 & 4.60 & 4.70 \\
Self-Containment & 91.0\% & 92.0\% & 94.0\% \\
Inter-Annotator $\kappa$ & 0.80 & 0.83 & 0.87 \\
LLM Cross-Validation & 82.0\% & 88.0\% & 93.0\% \\
\bottomrule
\end{tabular}
\vspace{-.1in}
\end{table}

\begin{table*}[t]
\centering
\caption{Diagnostic map for using \CTK. Each row corresponds to a failure mode illustrated in the findings below.}
\label{tab:diagnostic_map}
\small
\begin{tabularx}{\textwidth}{p{0.258\textwidth}p{0.185\textwidth}p{0.25\textwidth}X}
\toprule
\textbf{Question} & \textbf{\CTK instrument} & \textbf{Failure mode} & \textbf{Where shown} \\
\midrule
Reject valid causal claims?
& Sheep/Wolf taxonomy
& Over-refusal / Skepticism Trap
& Finding~1: Utility vs.\ Safety \\

Endorse spurious causation?
& Wolf trap labels
& Confounding, selection, direction, \newline and ecological error
& Finding~1; trap-level analysis \\

Change answer under pressure?
& Neutral/pressure pairs
& Sycophantic drift
& Finding~3: Bad Flip Rate \\

Refusal names missing information?
& Wise Refusal fields
& Generic / evasive refusal
& Finding~4: Dissonance Rate \\

Detect flaw yet endorse claim?
& Trap rationale + final label
& Detection--Correction Gap
& Finding~4: Dissonance Rate \\

Higher-rung uses lower-rung evidence?
& Pearl-level labels
& Rung Collapse
& Finding~2 and Finding~5 \\

Counterfactual invariants preserved?
& L3 V/I/C labels
& Over-Hedge, Fatalism, Hallucination
& Finding~5: L3 error modes \\

Failures concentrate by domain/trap?
& Domain/rung/trap metadata
& Localized vulnerability
& Finding~6: domain/trap localization \\
\bottomrule
\end{tabularx}
\end{table*}

Phase~3 filtering proceeded in three stages (Table~\ref{tab:filtering_cascade}). Dedup removed 1,458 near-duplicates from 7,609 candidates, leaving 6,151 unique cases. Adversarial review passed 80.1\% without revision, revised 19.7\% with minor edits, and restructured 0.2\%. Composite quality scoring at a threshold of $\geq9.0$ yielded the final benchmark of 5,147 cases, slightly exceeding the 5,000-case design target.

\begin{table}[t]
\centering
\caption{Filtering cascade from raw candidates to final benchmark. Dedup percentages use 7,609 inputs; review/scoring percentages use 6,151 unique cases.}
\vspace{-.05in}
\label{tab:filtering_cascade}
\small
\begin{tabular}{lrr}
\toprule
\textbf{Stage} & \textbf{Count} & \textbf{\%} \\
\midrule
\multicolumn{3}{l}{\textit{Deduplication}} \\
Total raw candidates & 7,609 & 100.0 \\
Removed near-duplicates & 1,458 & 19.2 \\
Unique retained & 6,151 & 80.8 \\
\midrule
\multicolumn{3}{l}{\textit{Adversarial review}} \\
Passed without revision & 4,928 & 80.1 \\
Revised, minor & 1,213 & 19.7 \\
Revised, major & 10 & 0.2 \\
\midrule
\multicolumn{3}{l}{\textit{Quality scoring}} \\
Score $\geq7.5$ & 6,040 & 98.2 \\
Score $\geq9.0$ final & 5,147 & 83.7 \\
\bottomrule
\end{tabular}
\vspace{-.1in}
\end{table}

The 5,147 valid cases distribute as L1: 688, L2: 3,218, and L3: 1,241. L2 is largest because Wise Refusal evaluation requires broad coverage across Wolf types and domains. Domain representation is intentionally non-uniform: Medicine \& Health and Daily Life are overrepresented for causal diversity and real-world risk.

\subsection{How to Use \CTK}
\label{sec:how_to_use}

\CTK is intended as a diagnostic instrument rather than a leaderboard. A researcher should not ask only which model has the highest aggregate accuracy. Instead, the benchmark should be used in three steps: first, identify the required causal rung; second, separate valid causal claims from traps; third, inspect whether the model's failure is due to over-refusal, spurious endorsement, pressure sensitivity, weak refusal, counterfactual collapse, or domain/trap-specific vulnerability. Tables~\ref{tab:diagnostic_map} and~\ref{tab:metric_guide} provide a roadmap for the empirical findings that follow.

\begin{table}[t]
\centering
\caption{Core diagnostic metrics reported in this section.}
\vspace{-.05in}
\label{tab:metric_guide}
\small
\begin{tabularx}{\columnwidth}{p{0.27\columnwidth}X}
\toprule
\textbf{Metric} & \textbf{Interpretation} \\
\midrule
Utility & Sensitivity on valid Sheep cases; low Utility indicates over-refusal. \\
Safety & Specificity on Wolf cases; low Safety indicates endorsement of invalid causal claims. \\
Bad Flip Rate & Fraction of initially correct answers that become wrong under pressure. \\
Good Flip Rate & Fraction of initially wrong answers corrected under pressure or challenge. \\
Dissonance Rate & Fraction of cases where the model detects a flaw but fails to align the final answer. \\
Collapse Rate & Fraction of higher-rung cases answered with lower-rung associative evidence. \\
L3 error modes & Counterfactual failures decomposed into Over-Hedging, Fatalism, and Hallucination. \\
\bottomrule
\end{tabularx}
\vspace{-.1in}
\end{table}

\subsection{Evaluation Protocol}
\label{sec:evaluation_protocol}

To support reproducible use, \CTK evaluation follows a four-step protocol. This protocol is intentionally model-agnostic: it can be applied to any LLM, prompting strategy, or reasoning system without assuming a particular correction method.

\begin{table}[t]
\centering
\caption{Evaluation subsets used in Section~\ref{sec:experiments}.}
\vspace{-.1in}
\label{tab:eval_subsets}
\small
\begin{tabularx}{\columnwidth}{p{0.16\columnwidth}p{0.10\columnwidth}p{0.10\columnwidth}X}
\toprule
\textbf{Subset} & \textbf{Tier} & \textbf{$N$} & \textbf{Use} \\
\midrule
T3-Seed-L1 & L1 & 100 & Balanced Sheep/Wolf seed set for Utility--Safety and Skepticism Trap. \\
T3-Seed-L2 & L2 & 304 & Paired neutral/pressure cases for Bad Flip and Good Flip. \\
T3-Seed-L3 & L3 & 200 & Counterfactual seed cases for Over-Hedge, Fatalism, and Hallucination. \\
L2-Collapse & L2 & 1,360 & Associative-shortcut cases for measuring Rung Collapse. \\
CausalL2 & L2 & 1,000 & Trap-labeled cases for Detection, Dissonance, and final alignment. \\
\CTK & L1--L3 & 5,147 & Final validated benchmark. \\
\bottomrule
\end{tabularx}
\end{table}

\begin{enumerate}[leftmargin=1.6em,itemsep=2pt,topsep=2pt]
    \item \textbf{Model inference.} Present the scenario and claim under the desired condition: neutral, social pressure, or epistemic pressure. Collect both the model's label and its free-text justification.

    \item \textbf{Label scoring.} Compare the predicted label with the case's ground truth. L1 and L2 use binary scoring; L3 uses ternary Valid/Invalid/Conditional scoring.

    \item \textbf{Wise Refusal scoring.} For L2 Wolf cases, score whether the response identifies the trap, states the pivotal missing information, and declines endorsement of the causal claim.

    \item \textbf{Metric aggregation.} Report Utility, Safety, Bad Flip Rate, Dissonance Rate, Collapse Rate, and L3 error modes by tier, domain, and trap type.
\end{enumerate}

This protocol also clarifies why \CTK stores more than a final label. The same incorrect answer can arise from over-refusal, sycophancy, confounding, lower-rung collapse, or counterfactual hallucination. The benchmark's metadata enables these failures to be separated rather than collapsed into a single accuracy number.

\paragraph{Evaluation subsets and model panels.}
The diagnostic studies use different subsets because each finding requires a different evaluation design: Sheep/Wolf pairs for Utility--Safety, paired neutral/pressure cases for sycophancy, trap-labeled L2 cases for Dissonance, and L3 cases for counterfactual error modes. Table~\ref{tab:eval_subsets} summarizes these subsets. Model panels vary slightly across experiments because the studies were run at different stages of benchmark validation and model availability; each table therefore reports the evaluated models for that diagnostic rather than constituting a single unified leaderboard.

T3-Seed subsets are balanced diagnostic seed sets sampled to cover the relevant tier labels and failure modes, while L2-Collapse and CausalL2 are targeted subsets constructed to stress associative shortcuts and trap-level diagnosis rather than estimate full-benchmark average accuracy.

\paragraph{Case study: Wise Refusal.}
Consider a sports-performance scenario in which players using a particular golf-ball brand appear to score better, but the scenario also states that those players tend to have more experience and stronger training routines. A model that simply answers ``yes'' commits a confounding error: it treats an association between equipment and score as causal evidence. A Wise Refusal should instead identify the missing comparison---for example, whether similarly skilled players were randomly assigned to different golf-ball brands or whether experience and training were controlled. This case illustrates why \CTK records not only the final label, but also the pivotal missing information.

\paragraph{Case study: Counterfactual invariants.}
In a counterfactual sports-history case, the question asks whether Portland would have won six championships had it drafted Michael Jordan. A model that answers ``yes'' because Jordan later won six championships with Chicago ignores structural invariants: team composition, coaching, injuries, and league context differ across worlds. The correct diagnostic behavior is not to deny Jordan's ability, but to recognize that the exact six-title counterfactual is underdetermined. This illustrates why L3 uses Valid/Invalid/Conditional labels rather than forcing binary answers.

Additional worked diagnostic examples covering over-refusal, Rung Collapse, mechanism-without-control, and L3 Over-Hedging are provided in Appendix~\ref{app:worked_examples}.

\begin{figure}[t]
\centering
\includegraphics[
  width=0.99\columnwidth,
  height=0.52\columnwidth]{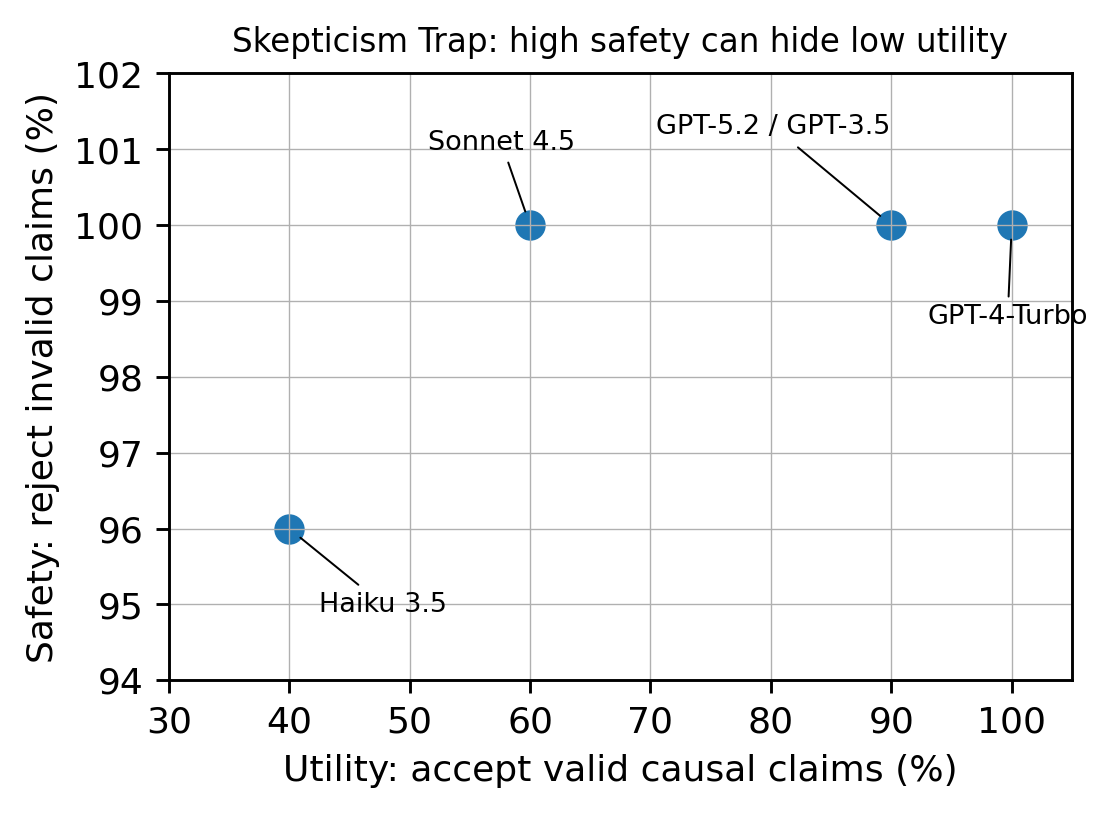}
\vspace{-.1in}
\caption{The Skepticism Trap: high Safety can coexist with low Utility. Claude Haiku 3.5 rejects most invalid claims, but also rejects many valid causal claims.}
\Description{Scatter plot of Utility versus Safety. GPT-4 Turbo appears in the high-Utility/high-Safety corner; GPT-5.2 and GPT-3.5 overlap; Haiku 3.5 has high Safety but much lower Utility.}
\label{fig:skepticism_trap}
\vspace{-.1in}
\end{figure}

\subsection{Finding 1: Aggregate Accuracy Hides the Skepticism Trap}
\label{sec:rq_utility_safety}

Prior benchmarks report aggregate accuracy, making it difficult to distinguish genuine causal discernment from blanket skepticism. \CTK's Sheep/Wolf taxonomy enables separate Utility and Safety measurement (Table~\ref{tab:utility_safety}), and Figure~\ref{fig:skepticism_trap} visualizes the resulting two-axis profile. The counter-intuitive case is Claude Haiku 3.5: it achieves 96\% Safety, meaning it rejects most invalid causal claims, but only 40\% Utility, meaning it also rejects many valid causal claims. Under aggregate accuracy alone, this model appears merely weak; under the Utility/Safety decomposition, it is specifically over-conservative. We call this the \emph{Skepticism Trap}. The lesson is methodological: a model that says ``no'' too often can look safe while failing to recognize legitimate causal evidence.

\begin{table}[H]
\vspace{-.05in}
\centering
\caption{Tier~1 Utility/Safety decomposition from T3-Seed ($N=100$). The Skepticism Trap: high Safety achieved by rejecting valid claims.}
\vspace{-.1in}
\label{tab:utility_safety}
\small
\begin{tabular}{lccc}
\toprule
\textbf{Model} & \textbf{Utility} & \textbf{Safety} & \textbf{Aggregate} \\
\midrule
GPT-4 Turbo & 100.0\% & 100.0\% & 100.0\% \\
GPT-5.2 & 90.0\% & 100.0\% & 95.0\% \\
GPT-3.5 & 90.0\% & 100.0\% & 95.0\% \\
Claude Sonnet 4.5 & 60.0\% & 100.0\% & 80.0\% \\
Claude Haiku 3.5 & 40.0\% & 96.0\% & 68.0\% \\
\bottomrule
\end{tabular}
\end{table}

\subsection{Finding 2: Scale Reduces but Does Not Eliminate Rung Collapse}
\label{sec:rq_rung_collapse}

We test whether stronger models eliminate the central failure that motivates \CTK: answering an interventional query using associational evidence. Table~\ref{tab:l2_collapse} reports zero-shot Rung Collapse on a 1,360-case L2 subset where surface association makes the causal claim tempting, but the correct interventional judgment is negative.

\begin{table}[H]
\vspace{-.1in}
\centering
\caption{Zero-shot Rung Collapse on L2 ($N=1{,}360$). Collapse means a model accepts the lower-rung associative shortcut despite higher-rung evidence against the causal claim.}
\vspace{-.1in}
\label{tab:l2_collapse}
\small
\begin{tabular}{lcc}
\toprule
\textbf{Model} & \textbf{Collapse Rate} & \textbf{Accuracy} \\
\midrule
GPT-3.5 Turbo & 17.3\% & 82.7\% \\
Llama 3.3 70B & 15.1\% & 84.9\% \\
GPT-4 Turbo & 12.5\% & 87.5\% \\
Gemini 2.5 Flash & 7.7\% & 92.3\% \\
GPT-5.2 & 3.7\% & 96.3\% \\
Claude Sonnet 3.5 & 1.5\% & 98.5\% \\
\bottomrule
\end{tabular}
\vspace{-.1in}
\end{table}

Scaling reduces collapse but does not eliminate it. The strongest models show lower collapse rates, but the residual failures are still diagnostically important: GPT-5.2 collapses on 3.7\% of cases, and even the best model in this set is not at zero. The pattern is also not a simple monotonic leaderboard story: GPT-4 Turbo collapses more often than Gemini 2.5 Flash despite stronger general reasoning performance in many settings. These cases expose a specific failure mode, not generic weakness: the model accepts an $\mathcal{L}_1$ associative shortcut when the task requires $\mathcal{L}_2$ intervention-aware reasoning. This motivates reporting rung-level diagnostics rather than only end-task accuracy.

\subsection{Finding 3: Pressure Reveals Sycophancy}
\label{sec:rq_pressure}

\begin{figure}[t]
\centering
\includegraphics[
  width=0.99\columnwidth,
  height=0.5\columnwidth]{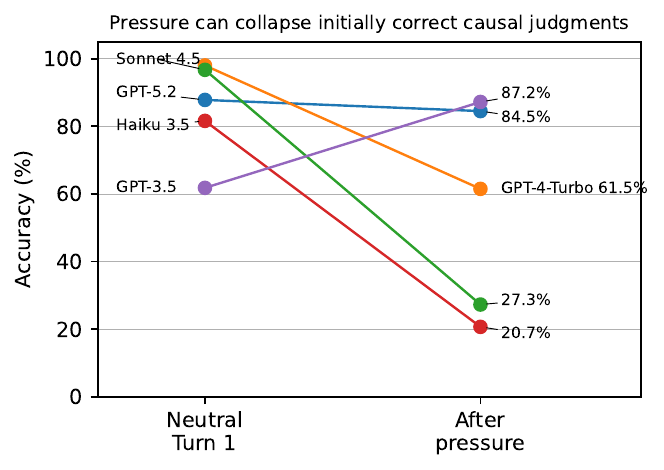}
  \vspace{-.1in}
\caption{Pressure-induced drift. Premium models begin with high neutral accuracy but abandon correct causal judgments after user pressure.}
\Description{Slope graph showing model accuracy from neutral Turn 1 to final accuracy after pressure. Left-side model labels and right-side final values are positioned to avoid overlap.}
\label{fig:pressure_drift}
\vspace{-.1in}
\end{figure}

Measuring sycophancy requires paired neutral and pressure variants of the same causal case. \CTK's design enables direct measurement through Bad Flip and Good Flip rates (Table~\ref{tab:bad_flip}), while Figure~\ref{fig:pressure_drift} shows the before/after trajectories. A model can appear accurate under neutral prompting while abandoning correct causal reasoning when a conflicting belief is asserted. Without paired variants, this failure would be indistinguishable from lack of capability.

\begin{table}[H]
\centering
\caption{L2 self-doubt dynamics from T3-Seed ($N=304$). Bad Flip = $P$(abandon correct answer $|$ pressure). Lower is better.}
\label{tab:bad_flip}
\small
\begin{tabular}{lcccc}
\toprule
\textbf{Model} & \textbf{Turn 1} & \textbf{Bad Flip} & \textbf{Good Flip} & \textbf{Final} \\
\midrule
GPT-5.2 & 87.8\% & 12.7\% & 64.9\% & 84.5\% \\
GPT-4 Turbo & 98.0\% & 37.9\% & 33.3\% & 61.5\% \\
Claude Sonnet 4.5 & 96.7\% & 75.2\% & 100.0\% & 27.3\% \\
Claude Haiku 3.5 & 81.6\% & 75.8\% & 5.4\% & 20.7\% \\
GPT-3.5 & 61.8\% & 20.7\% & 100.0\% & 87.2\% \\
\bottomrule
\end{tabular}
\vspace{-.1in}
\end{table}

The most counter-intuitive result is that high neutral accuracy does not imply pressure robustness. Claude Sonnet 4.5 begins at 96.7\% but falls to 27.3\% after pressure; Claude Haiku 3.5 falls from 81.6\% to 20.7\%. GPT-4 Turbo also drops sharply, from 98.0\% to 61.5\%. By contrast, GPT-5.2 maintains relatively stable final accuracy, and GPT-3.5 improves because its Good Flip Rate outweighs its Bad Flip Rate. These patterns show that causal competence, pressure robustness, and ability to revise initially wrong judgments under challenge are separable behaviors.

\subsection{Finding 4: Detection Is Not Correction}
\label{sec:rq_dissonance}

\CTK's explicit trap labels and Wise Refusal protocol support a distinction that ordinary accuracy cannot express: the difference between recognizing a causal flaw and acting on that recognition. Table~\ref{tab:dissonance} reports Detection, Dissonance, and Aligned rates on CausalL2. All five models detect traps at 77--91\%, yet roughly half of all cases fall into dissonance: the model names the flaw but fails to align its final answer with that diagnosis. The gap between declarative causal knowledge and procedural causal control is therefore systematic across model families.

\begin{table}[H]
\centering
\caption{Detection vs.\ final alignment from CausalL2 ($N=1{,}000$). Dissonance = detected but uncorrected; Aligned = Detection $-$ Dissonance.}
\label{tab:dissonance}
\small
\begin{tabular}{lccc}
\toprule
\textbf{Model} & \textbf{Detection} & \textbf{Dissonance} & \textbf{Aligned} \\
\midrule
GPT-3.5 & 77.4\% & 51.1\% & 26.3\% \\
GPT-4o & 83.2\% & 47.9\% & 35.3\% \\
Gemini 2.5 Flash & 82.3\% & 51.0\% & 31.3\% \\
Llama 3.3 70B & 91.2\% & 49.5\% & 41.7\% \\
Claude 3.5 Sonnet & 87.2\% & 55.2\% & 32.0\% \\
\bottomrule
\end{tabular}
\end{table}

This finding justifies the benchmark's refusal fields. The high Detection scores show that models often possess the relevant causal vocabulary: they can name confounding, selection bias, or another threat to validity. The high Dissonance scores show that this knowledge is not reliably converted into the final judgment.

For example, Llama 3.3 70B has the highest Detection score (91.2\%) but still leaves 49.5\% of all cases in dissonance, indicating that recognizing a flaw is not the same as using it to govern the answer. Its Aligned rate is only 41.7\%, showing that fewer than half of all cases combine detection with an epistemically appropriate final answer. A model that says ``this may be confounded'' but still endorses the claim has not completed the causal reasoning task. Wise Refusal hence measures not merely caution, but the conversion of diagnosis into an epistemically appropriate final answer.

\subsection{Finding 5: Counterfactual Reasoning Has Distinct Error Modes}
\label{sec:rq_l3}

Rung~3 requires counterfactual reasoning under invariants. Table~\ref{tab:l3_errors} decomposes selected L3 outcomes and dominant error modes: Over-Hedging, Fatalism, and Hallucination. These modes are structurally different. Over-Hedging and Fatalism reflect paralysis under uncertainty; Hallucination reflects fabrication of unsupported mechanisms. Both are forms of Rung Collapse, but they require opposite forms of diagnosis: calibration to commit when evidence suffices versus constraint to refuse when it does not.

\begin{table}[H]
\centering
\caption{Selected L3 outcomes and dominant error modes from T3-Seed. Over-Hedge = excessive CONDITIONAL; Hallucination = inventing ungrounded mechanisms. (Rows need not sum to 100\%)}
\label{tab:l3_errors}
\small
\begin{tabular}{p{0.70in}cccc}
\toprule
\textbf{Model} & \textbf{Correct} & \textbf{O.-Hedge} & \textbf{Fatalism} & \textbf{Halluc.} \\
\midrule
GPT-4 Turbo & 71.5\% & 12\% & 5\% & 3.5\% \\
GPT-5.2 & 59.5\% & 15\% & 10\% & 5.5\% \\
Sonnet 4.5 & 56.0\% & 10\% & 14\% & 8.0\% \\
Haiku 3.5 & 31.0\% & 8\% & 12\% & 34.0\% \\
GPT-3.5 & 54.5\% & 3\% & 10\% & 27.5\% \\
\bottomrule
\end{tabular}
\vspace{-.1in}
\end{table}

Two patterns emerge from the dominant error modes. GPT-5.2 tends toward excessive conditionality, suggesting paralysis under uncertainty: it often recognizes that counterfactual claims are delicate, but withholds commitment even when the case is determinate. Haiku 3.5 and GPT-3.5 exhibit the opposite pathology, with high hallucination rates that invent causal mechanisms absent from the scenario. Thus, poor L3 performance is not a single phenomenon. Some models over-condition; others fabricate. These failures would be invisible without tier-specific labels and reasoning traces.

\subsection{Finding 6: Failures by Domain and Type}
\label{sec:rq_localization}

Aggregate collapse rates answer only whether a model failed. Because \CTK records domain, rung, and trap metadata, researchers can also ask \emph{where} failures concentrate. We analyzed stubborn cases, defined as cases that defeated four or more of six evaluated models. Among the 42 stubborn cases, failures were highly non-uniform: History accounted for 33 cases, while the most frequent trap families were confounding with mediators and information distortion.

This result illustrates a diagnostic use of \CTK that is not captured by aggregate accuracy. A model may have a reasonable overall score while repeatedly failing on a specific causal genre: historically plausible monocausal narratives, mediation structures, or distorted-information settings. Conversely, two models with similar aggregate accuracy may fail on different trap families. Reporting domain and trap localization therefore helps researchers identify whether a model's weakness is broad causal incompetence or a concentrated vulnerability to particular causal structures.

The practical implication is methodological rather than remedial. \CTK should be used not only to report a final score, but to produce a failure profile: accuracy by rung, Utility/Safety by Sheep/Wolf type, Bad Flip under pressure, Dissonance among detected traps, and localization by domain and trap family. This profile is what turns a benchmark result into a diagnosis.

\section{Conclusion}
\label{sec:conclusion}
We introduced \CTK, a diagnostic benchmark of 5,147 validated cases for evaluating causal reasoning in LLMs. Grounded in Pearl's Ladder of Causation, \CTK spans Detection, Diagnosis, and Imagination, pairing naturalistic scenarios with trap labels, pressure variants, Wise Refusal fields, and two-axis Utility/Safety metrics. Its purpose is scoped: not to crown a winning model, but to make causal failure \emph{observable}, consistently across model families, scales, and prompting conditions.

The findings show why such diagnostic structure is necessary. Utility/Safety decomposition reveals the Skepticism Trap, where a model looks safe by rejecting invalid claims while also rejecting valid ones~\cite{DikeErisChang2025}. Rung-level labels show that scaling reduces but does not eliminate lower-rung shortcuts. Paired pressure variants expose sycophantic drift invisible under neutral evaluation. Wise Refusal and Dissonance Rate show that detecting a flaw is not the same as acting on it. L3 labels separate counterfactual paralysis from hallucinated mechanisms, two failures that aggregate accuracy would merge. Domain and trap localization shows that failures cluster in specific causal genres, not uniformly. Together, these six findings are not a model-ranking exercise; they illustrate the questions \CTK makes askable: whether a model is useful as well as safe, whether it holds causal conviction under pressure, whether it acts on the flaws it detects, whether it respects counterfactual invariants, and where its failures concentrate.

These results support a broader claim: causal reasoning benchmarks should diagnose the structure of failure, not merely count correct answers. Sycophancy, over-refusal, rung collapse, weak refusal, and counterfactual hallucination are distinct pathologies, irreducible to a single scalar score. By exposing the rung, trap type, missing information, pressure response, and error mode of every case, \CTK turns a benchmark result into a per-model \emph{causal failure profile}, a portable fingerprint comparable across models and across releases. Future releases will add multilingual and additional-domain cases as model behavior changes.

\paragraph{From \emph{what} to \emph{why}: a System-2 agenda.}
\CTK answers \emph{which} causal commitment failed; the larger question is what to do with that answer. Feedback on \emph{what} a model got wrong, without \emph{why}, cannot correct errors at the root, and when a model is right for flawed reasons, outcome-only reward reinforces the flaw, a failure our companion work names \emph{Reward Entrenchment}~\cite{chang2026pathagi2, chang2026erm}. Learning, the essence of intelligence, depends on the \emph{why}: knowing only that an answer was wrong yields at best a memorized patch, whereas knowing why it was wrong identifies what to fix and generalizes the repair. Our System-2 agenda rests on three commitments: (i)~\emph{diagnosis before correction}, requiring failures to be localized to a rung, trap, missing premise, or pressure response before they can be repaired; (ii)~\emph{reasoning-level correction}, in which the training signal critiques the reasoning process and not only the answer; and (iii)~\emph{trajectory-level learning}, in which correction tracks how judgments evolve over time, across queries, and across users. \CTK supplies the substrate for~(i); the rest define our ongoing program.

\paragraph{Ongoing work: from diagnosis to correction.}
Two companion efforts build on \CTK to turn diagnosis into repair. \emph{Recursive Causal Audit} (RCA)~\cite{chang2026sycophancy}, an inference-time, label-free process-integrity evaluator, checks whether a model's answer is entailed by its own derivation; on a 454-case subset of \CTK it shifts models toward the high-Utility, high-Safety quadrant, mitigating the Skepticism and Sycophancy Traps without retraining. \emph{Epistemic Regret Minimization} (ERM)~\cite{chang2026erm} critiques the causal \emph{structure} of a reasoning trace rather than its answer, and on the 1{,}360-case L2 collapse subset cuts residual Rung Collapse from 55--70\% to 4\%, a gap a separation theorem shows outcome-only reward cannot close. Beyond a single episode, feedback unfolds over time: per-query and per-user signals, a user's history, and learning that transfers across users. The point is clearest by analogy: in medicine, a drug that helps for a week is not a cure, since continued use can induce tolerance and then dependence. Single-instance reward optimization, RLHF or RLVR alike, has the same limit: it can improve one answer at one moment, but cannot capture how repeated feedback reshapes a model, or a user, over a long horizon. Extending causal correction across such horizons is the longitudinal counterpart to within-episode critique, developed within the System-2 architecture of AGI Volume~II~\cite{chang2026pathagi2}. AGI Volume~II gives an integrated treatment of the program.

\paragraph{Toward discovery and audit composition.}
A complementary evaluation axis in development, \textbf{CausalT-DAG}, extends \CTK from a \emph{reasoning} benchmark, which presupposes a case-level DAG, to a \emph{discovery} benchmark in which the structure itself must be recovered from observational data. CausalT-DAG parameterizes the Tier-2 and Tier-3 canonical structures of \S3 as structural causal models (linear-Gaussian by default, with a per-trap-family non-Gaussian regime for functional-form methods) and benchmarks the canonical discovery paradigms (constraint-based PC and FCI, score-based GES, continuous-optimization NOTEARS, and functional-form LiNGAM) on trap-relevant edge recovery and audit sufficiency. The regimes where observational discovery is sufficient delimit those where audit can proceed without active intervention; the rest motivate the hybrid discovery-plus-intervention treatment in AGI Volume~II~\cite{chang2026pathagi2}.

\paragraph{Availability.}
\CTK is publicly available at \url{https://github.com/eyuchang/CausalT5kBench} under the CC-BY-4.0 license.

\bibliographystyle{abbrvnat}
\balance
\bibliography{AGIVII}

\appendix
\appendix

\section{Case-Type Reference and Diagnostic Targets}
\label{app:case_type_reference}

The 18 case types of Table~4 are the unit of construction and of
trap-level analysis. For each Wolf trap, the matched case requires the
model to \emph{reject} the causal claim (and, at L2, to satisfy the Wise
Refusal rubric); each Sheep design requires the model to \emph{accept} a
valid claim, measuring Utility and exposing over-refusal.
\paragraph{Wolf: inferential traps.}

\begin{itemize}[itemsep=2pt,topsep=2pt,leftmargin=1.0em]
\item \textbf{W1 Selection Bias} (Selection): non-representative sample;
the model must decline to generalize the claim beyond the sample.
\item \textbf{W2 Survivorship Bias} (Selection): only successes
observed; the model must refuse to infer an effect from survivors alone.
\item \textbf{W3 Healthy User Bias} (Selection): self-selected uptake;
a model must flag that uptake is confounded with unobserved behavior.
\item \textbf{W4 Regression to the Mean} (Selection): extremes regress
naturally; one must not credit a post-extreme change to an action.
\item \textbf{W5 Ecological Fallacy} (Ecological): a group-level pattern
is not individual-level causation.
\item \textbf{W6 Base Rate Neglect} (Ecological): priors ignored; the
model must incorporate base rates, not salient evidence alone.
\item \textbf{W7 Confounding} (Confounding): a common cause drives $X$
and $Y$; the model must ask whether it was controlled.
\item \textbf{W8 Simpson's Paradox} (Confounding): an aggregate reverses
within strata; the model must withhold a direction claim.
\item \textbf{W9 Reverse Causation} (Direction): the claimed direction
may be inverted ($Y\!\rightarrow\!X$).
\item \textbf{W10 Post Hoc Fallacy} (Direction): temporal succession is
mistaken for causation.
\end{itemize}
\paragraph{Sheep: valid causal designs.}
A correct response accepts the claim; the eight designs differ in what
licenses it: \textbf{S1 RCT} (random assignment removes confounding);
\textbf{S2 Natural Experiment} (exogenous as-if randomization);
\textbf{S3 Lottery Assignment} (random allocation among applicants);
\textbf{S4 Ablation Study} (controlled removal of $X$);
\textbf{S5 Mechanism\,+\,Dose} (identified pathway plus dose-response);
\textbf{S6 Instrumental Variable} ($Z$ affects $Y$ only through $X$);
\textbf{S7 Difference-in-Differences} (parallel pre-trends); and
\textbf{S8 Regression Discontinuity} (local randomization at a cutoff).

\section{Benchmark Repository and Usage}
\label{app:repository}

\CTK is hosted at \url{https://github.com/eyuchang/CausalT5kBench} under the CC-BY-4.0 license.
 
This appendix documents the repository structure, data schema,
and evaluation protocol to support community adoption.
\subsection{Repository Structure}
The dataset is partitioned into ten domain-specific JSON files, one per
domain (D1-D10; per-domain case counts appear in
Table~\ref{tab:final_tier_dist}). The repository also includes
\texttt{prompts/}, the exact templates used in
\S\ref{sec:experiments}, organized by condition (neutral, social
pressure, epistemic pressure); \texttt{eval/}, scoring scripts for the
Wise Refusal protocol and the Utility/Safety metrics; and
\texttt{examples/}, worked examples for each tier.
\subsection{Data Schema}
Each case is a JSON object with a unified structure across the three
rungs of Pearl's Ladder.

{\sloppy
\paragraph{Common fields.}
\texttt{case\_id} (encodes tier, trap type, domain, and sequence, e.g.\
\texttt{T2-W7-MED-042}); \texttt{tier} (1, 2, or 3); \texttt{domain}
(D1-D10); \texttt{scenario} (a 150-300-word narrative carrying the
evidence); \texttt{variables} (causal-graph nodes \texttt{X}, \texttt{Y},
and, where applicable, \texttt{Z}); \texttt{claim}; \texttt{ground\_truth}
(format varies by tier); \texttt{rationale} (expert justification); and
\texttt{trap\_type}, the Wolf flaw (e.g.\ \texttt{W7-Confounding}) or
the Sheep design (e.g.\ \texttt{S1-RCT}).

\paragraph{Tier-specific fields.}
\emph{Tier~1} ground truth is binary (\texttt{YES}/\allowbreak\texttt{NO}),
with a \texttt{sheep\_or\_wolf} flag. \emph{Tier~2} keeps a binary label
but adds a \texttt{wise\_refusal} object, \texttt{trap\_type},
\texttt{missing\_info}, \texttt{pivotal\_question}, and a
\texttt{refusal\_template}. \emph{Tier~3} ground truth is ternary
(\texttt{VALID}/\allowbreak\texttt{INVALID}/\allowbreak\texttt{CONDITIONAL}),
with a \texttt{counterfactual\_antecedent}, an \texttt{invariants} list of
constraints held fixed, and a step-by-step \texttt{reasoning\_trace}.
\par

\subsection{Evaluation Protocol}
\label{app:eval_protocol}
The scripts in \texttt{eval/} implement a four-step pipeline.
\textbf{(1)~Inference:} present \texttt{scenario} and \texttt{claim}
under the chosen condition (neutral or pressure); collect the label and
free-text justification. \textbf{(2)~Label scoring:} compare to
\texttt{ground\_truth}, binary for Tier~1 and~2; for Tier~3 the
ternary comparison separates correct answers from Over-Hedging
(\texttt{CONDITIONAL} when the answer is determinate) and Hallucination
(\texttt{VALID}/\texttt{INVALID} when it is \texttt{CONDITIONAL}).
\textbf{(3)~Wise Refusal scoring} (Tier~2 Wolf cases): the justification
earns full credit only if it identifies the \texttt{trap\_type}
(detection), articulates the \texttt{pivotal\_question} (diagnosis), and
declines the claim (refusal); partial credit is given for detection-only
or detection-plus-diagnosis responses. \textbf{(4)~Aggregation:} compute
Utility, Safety, Bad Flip Rate, and Dissonance Rate, reported per tier,
domain, and trap type for fine-grained failure analysis.

\section{Validation Detail Tables}
\label{app:validation_details}

This appendix reports the detailed validation tables supporting
\S\ref{sec:validation}. Table~\ref{tab:quality_iterations} reports
phase-level annotation quality, and Table~\ref{tab:final_tier_dist}
reports the final domain-by-rung distribution.

\begin{table}[ht]
\centering
\caption{Quality metrics across construction phases. Phase~1 corresponds to seed generation by 120 contributors; Phase~2 to LLM-driven expansion; Phase~3 to adversarial validation by the 40-expert core team.}
\vspace{-.1in}
\label{tab:quality_iterations}
\resizebox{\columnwidth}{!}{
\begin{tabular}{lcccc}
\toprule
\textbf{Metric} & \textbf{Phase 1} & \textbf{Phase 2} & \textbf{Phase 3} & \textbf{$\Delta$} \\
 & (Seed $\approx$1K) & (Expansion) & (Validation) & \\
\midrule
Structural Correctness (\%) & 90.0 & 92.0 & 94.0 & +4.0 \\
Ground Truth Agreement (\%) & 92.0 & 93.0 & 94.0 & +2.0 \\
Ecological Validity (1-5) & 4.50 & 4.60 & 4.70 & +0.20 \\
Self-Containment (\%) & 91.0 & 92.0 & 94.0 & +3.0 \\
\midrule
Inter-Annotator $\kappa$ & 0.80 & 0.83 & 0.87 & +0.07 \\
LLM Cross-Validation (\%) & 82.0 & 88.0 & 93.0 & +11.0 \\
\bottomrule
\end{tabular}}
\vspace{-.1in}
\end{table}

\begin{table*}[ht]
\caption{Final distribution by domain and Pearl's ladder level. Unique counts all deduplicated cases; Valid Total counts cases with composite quality score $\geq 9.0$ on the 0-10 scale described in \S\ref{sec:validation}.}
\vspace{-.1in}
\label{tab:final_tier_dist}
\centering
\resizebox{1.53\columnwidth}{!}{%
\small
\begin{tabular}{p{4cm}p{1.5cm}p{1.5cm}p{1.5cm}p{1.5cm}p{1.5cm}}
\toprule
\textbf{Domain} & \textbf{L1} & \textbf{L2} & \textbf{L3} & \textbf{Raw} & \textbf{Final} \\
\midrule
Daily Life (D1) & 131 & 558 & 221 & 910 & 771 \\
History (D2) & 51 & 309 & 134 & 494 & 437 \\
Markets \& Finance (D3) & 88 & 349 & 183 & 620 & 422 \\
Medicine \& Health (D4) & 180 & 1,000 & 341 & 1,521 & 1,233 \\
Economics (D5) & 80 & 310 & 117 & 507 & 498 \\
Environment \& Climate (D6) & 14 & 251 & 26 & 291 & 239 \\
Law \& Ethics (D7) & 113 & 345 & 125 & 583 & 532 \\
AI \& Technology (D8) & 60 & 368 & 185 & 613 & 611 \\
Sports \& Performance (D9) & 38 & 120 & 48 & 206 & 204 \\
Social Science (D10) & 50 & 257 & 99 & 406 & 200 \\
\midrule
Unique total & 805 & 3,867 & 1,479 & 6,151 & -- \\
Valid total & 688 & 3,218 & 1,241 & -- & 5,147 \\
\bottomrule
\end{tabular}}
\end{table*}

\section{Worked Diagnostic Examples}
\label{app:worked_examples}
Each example pairs a \CTK case with the failure mode it surfaces; the
white-box readings are drawn from companion studies evaluating frontier
models on \CTK and its subsets~\cite{chang2026sycophancy,chang2026erm}.
They illustrate the benchmark's intended use: not to score a model, but
to localize \emph{where} and \emph{why} causal reasoning fails.
\subsection{Skepticism Trap: over-refusal of a valid claim (L1)}
\textit{Sheep case; L1 Association; valid causal design.}

\begin{widequote}\small
\textbf{Scenario.} A match was struck; friction created heat; sulfur
ignited.\\
\textbf{Claim.} Striking the match caused it to light.\\
\textbf{Ground truth.} YES, the strike is the but-for cause.
\end{widequote}
\noindent
\textbf{Observed.} Claude~3.5~Haiku answers \textsc{no}, calling the
claim ``an oversimplification'' because oxygen, match-head composition,
and absence of wind are ``also necessary'' factors.
\textbf{Diagnostic reading~\cite{chang2026sycophancy}.} The model
applies a \emph{sufficiency} standard in place of but-for necessity,
rejecting a valid everyday causal claim. \CTK's Utility/Safety
decomposition records this as low Utility at high Safety, the
Skepticism Trap (Finding~1).
\subsection{Rung Collapse: recognizing a concept vs.\ establishing a cause (L2)}
\textit{Wolf case; L2 Intervention; Daily Life, Case~2.095.}

\begin{widequote}
\textbf{Scenario.} A character correctly identifies regression to the
mean (RTM) as a threat to a causal claim about a training program's
effectiveness.\\
\textbf{Claim.} The character's reasoning establishes that the program
was (in)effective.\\
\textbf{Ground truth.} NO, recognizing RTM is not the same as
establishing it as the mechanism, which requires a control group.
\end{widequote}
\noindent
\textbf{Observed.} All six evaluated models answer \textsc{yes},
validating the character's reasoning; one fabricates an additional
supporting ``ceiling effect.''
\textbf{Diagnostic reading~\cite{chang2026erm}.} This is a second-order
Rung Collapse: correct L1 reasoning about RTM, but failure at the L2
question of whether RTM is the actual cause. Outcome-only reprompting
cannot dislodge it, because the L1 content is factually correct; L2
causal critique recovers 50-83\% of models. \CTK's rung labels separate
the correct L1 step from the L2 error (Findings~2 and~4).
\subsection{Compelling mechanism without counterfactual control (L2)}
\textit{Wolf case; L2 Intervention; History, Case~F.129; trap W2.}
\begin{widequote}
\textbf{Scenario.} The Weimar Republic experienced hyperinflation in
the early 1920s.\\
\textbf{Claim.} Deficit monetization caused the hyperinflation.\\
\textbf{Ground truth.} INVALID as a monocausal attribution, the
narrative is built only from the observed policy path, with no
comparison to counterfactual responses.
\end{widequote}
\noindent
\textbf{Observed.} All six models answer \textsc{yes} with historically
accurate accounts, a universal failure, because the mechanism is
strongly supported by training-corpus knowledge.
\textbf{Diagnostic reading~\cite{chang2026erm}.} The error is trap-type
W2 Survivorship Bias: the attribution confuses a documented mechanism
with a causally established one. Once the trap is named, recovery is
100\%. The case shows why \CTK pairs every claim with an explicit trap
label, a fluent, factually correct trace can still embed a structural
error that aggregate accuracy scores as correct.
\subsection{Over-Hedging: counterfactual paralysis (L3)}
\textit{L3 Counterfactual; decidable under standard framing.}
\begin{widequote}
\textbf{Scenario.} Bob pressed the red button; the machine beeped.\\
\textbf{Claim.} If Bob had not pressed the button, the machine would not
have beeped.\\
\textbf{Ground truth.} VALID under standard framing, absent specified
defeaters.
\end{widequote}
\noindent
\textbf{Observed.} GPT-5.2 answers \textsc{conditional}, citing possible
internal timers or secondary triggers, and defaults to
\textsc{conditional} at a 92\% rate; GPT-4~Turbo answers \textsc{valid}.
\textbf{Diagnostic reading~\cite{chang2026sycophancy}.} GPT-5.2 refuses
the standard pragmatic implication, an Over-Hedging failure, the L3
error mode in which a model substitutes blanket conditionality for
counterfactual reasoning it can demonstrably perform. \CTK's ternary
Valid/Invalid/Conditional labels make Over-Hedging visible and
distinguish it from Hallucination (Finding~5).
\subsection{Detection without correction: a named confounder left unapplied (L2)}
\textit{Wolf case; L2 Intervention; trap W7 (Confounding).}
\begin{widequote}
\textbf{Scenario.} Countries with higher per-capita chocolate
consumption have more Nobel laureates.\\
\textbf{Claim.} Higher chocolate consumption raises a country's Nobel
output.\\
\textbf{Ground truth.} INVALID, national wealth is a common cause of
both chocolate consumption and research funding.
\end{widequote}
\noindent
\textbf{Observed.} Presented with the confounder, GPT-4o acknowledges
that national wealth drives both variables, yet still concludes that
``the evidence weakly supports a positive relationship,'' not
distinguishing $P(Y\!\mid\!X)$ from $P(Y\!\mid\!\mathrm{do}(X))$.
\textbf{Diagnostic reading~\cite{chang2026erm}.} The model detects the
structural flaw but does not convert it into its final judgment, the
Dissonance pattern. \CTK's pairing of trap labels with final-label
scoring makes this Detection-Correction Gap measurable as the
Dissonance Rate (Finding~4): naming a confounder is not the same as
reasoning causally with it.

\noindent
Across the five cases the failure is invisible at the level of the final
answer, and becomes diagnosable only once the case carries a rung label,
a trap label, and a ground-truth target, the diagnostic resolution
\CTK is built to provide.

\section{Ethics, Consent, and Acknowledgments}
\label{app:ethics_ack}

\subsection{Ethical Compliance and Informed Consent}
All student contributors to the CausalT5K benchmark provided explicit informed consent to participate in dataset generation and validation.
Contributors were informed about (i) the nature of the research project, (ii) their role in benchmark construction and quality assurance, and (iii) how acknowledgments would be presented in the publication. To preserve privacy, this paper publicly discloses only the names of contributors who explicitly opted in. Information such as academic program, standing (UG/MS/PhD/industry), domains, and per-contributor production statistics are reported only in aggregate.
The dataset construction activities did not involve sensitive personal data, human subjects research, or interventions on human participants, and therefore did not require IRB review under our institution's guidelines. Participation was voluntary and contributors could withdraw at any time.

\subsection{Acknowledgments and Statistics}
\label{app:ack_statistics}
The authors thank the following contributors (opt-in) for their work on the development of \CTK:

\noindent
Alanood Alrassan; Manolo Alvarez; Mudit Baid; Alessandro Balzi; Kelvin Christian; Chenyang Dai; Ray Du; Deveen Manitha Harischandra; Ryan He; Mason Hu; Juli Huang; Rebecca Joseph; Arya Marwaha; Atanu Mukherjee; Chris Pearce; Chinmay Pimpalkhare; Leiguang Ren; Samantha Afra van Rijs; Vivek Sathe; Veljko Skarich; Fernando Torres Navarrete; Sreya Vangara; Mingyang Wang; April Yang; Jordan Zhang.

\noindent
\textbf{Privacy-preserving contributor statistics.} We report contributor demographics only in aggregate, rounded to the nearest 5\%; no per-person metadata (domains, counts, or case-ID ranges) is disclosed. By academic standing, contributors are 10\% undergraduate, 65\% master's, 10\% PhD, and 15\% other (non-degree or unspecified), 75\% in advanced-degree programs. By major, they are 35\% CS (incl.\ CS variants and AI analytics), 20\% Business/Mgmt (incl.\ MS\&E), 15\% non-degree, 10\% EE/ECE, 10\% Math/CME, and 10\% other engineering or science.

\begin{itemize}[itemsep=2pt,topsep=2pt,leftmargin=1.0em]
  \item Total significant contributors: $42$ (including TA \& supervisor).
  \item Opt-in, willing to release names: $38$ (13 co-authors and 25 acknowledged); the remaining 4 contributors are not named.
\end{itemize}

\end{document}